\documentclass[10pt,twocolumn]{IEEEtran}
\usepackage{times}
\usepackage{epsfig}
\usepackage{graphicx}
\usepackage{amsmath}
\usepackage{amssymb}
\usepackage{times}
\usepackage[utf8]{inputenc}
\usepackage{enumitem}
\usepackage{epsfig}
\usepackage{graphicx}
\usepackage{amsmath}
\usepackage{amssymb}
\usepackage{algorithm}
\usepackage{algorithmic}
\usepackage{subcaption}
\usepackage{multirow}
\usepackage{makecell}

\usepackage{xcolor}
\usepackage{color}
\usepackage{comment}
\usepackage{cuted}
\usepackage{capt-of}
\usepackage{graphicx}
\usepackage{times}
\usepackage{epsfig}
\usepackage{graphicx}
\usepackage{amsmath}
\usepackage{amssymb}
\usepackage{booktabs}

\definecolor{aj}{rgb}{0.8, 0.33, 0.0}

\definecolor{ms}{rgb}{0.1, 0.5, 0.5}
\newcommand{\MS}[1]{{\color{ms} Mubarak: #1}}

\usepackage[pagebackref=true,breaklinks=true,colorlinks,bookmarks=false]{hyperref}

\usepackage[breaklinks=true,bookmarks=false]{hyperref}

\begin{document}

%%%%%%%%% TITLE
\title{LSDAT: Low-Rank and Sparse Decomposition for Decision-based Adversarial Attack}

\author{
    \IEEEauthorblockN{Author1\IEEEauthorrefmark{1}, Author2\IEEEauthorrefmark{2}, Author3\IEEEauthorrefmark{2}, Author4\IEEEauthorrefmark{1}}
    \IEEEauthorblockA{\IEEEauthorrefmark{1}Institution1
    \\\{1, 4\}@abc.com}
    \IEEEauthorblockA{\IEEEauthorrefmark{2}Institution2
    \\\{2, 3\}@def.com}
}

\author{Ashkan Esmaeili\textsuperscript{1} \IEEEauthorrefmark{1}, Marzieh Edraki\textsuperscript{1} \IEEEauthorrefmark{1}, Nazanin Rahnavard \IEEEauthorrefmark{1}, Mubarak Shah \IEEEauthorrefmark{2}, Ajmal Mian \IEEEauthorrefmark{3}

\IEEEauthorblockA{\IEEEauthorrefmark{1} University of Central Florida} %
\IEEEauthorblockA{\IEEEauthorrefmark{2} Center for Research in Computer Vision}                      
                      \IEEEauthorblockA{\IEEEauthorrefmark{3} University of Western Australia}
%{\tt\small ashkan.esmaeili@ucf.edu}
% \and
{\tt\small ashkan.esmaeili@ucf.edu}{\quad \quad  \small{1 indicates shared first authorship}}
}
\maketitle
%\ificcvfinal\thispagestyle{empty}\fi
%%%%%%%%% ABSTRACT

\begin{abstract}
%%%%%%%%% ABSTRACT
We propose LSDAT, an image-agnostic 
decision-based black-box attack that exploits low-rank and sparse decomposition (LSD) to dramatically reduce the number of queries and achieve superior fooling rates compared to the state-of-the-art decision-based methods under given imperceptibility constraints. 
LSDAT crafts perturbations in the low-dimensional subspace formed by the sparse component of the input sample and that of an 
adversarial sample to obtain query-efficiency. The specific perturbation of interest is obtained by traversing the path between the input and adversarial sparse components. It is set forth that the proposed sparse perturbation is the most aligned sparse perturbation with the shortest path from the input sample to the decision boundary for some initial adversarial sample (the best sparse approximation of shortest path, likely to fool the model). 
Theoretical analyses are provided to justify the functionality of LSDAT. Unlike other dimensionality reduction based techniques  
aimed at improving query efficiency (e.g, ones based on FFT), LSD works directly in the image pixel domain to guarantee that non-$\ell_2$ constraints, such as sparsity, are satisfied. LSD offers better control over the number of queries and provides computational efficiency as it performs sparse decomposition of the input and 
adversarial images only once to generate all queries. 
We demonstrate $\ell_0$, $\ell_2$ and $\ell_\infty$ bounded attacks with LSDAT to evince its efficiency compared to baseline decision-based attacks in diverse low-query budget scenarios as outlined in the experiments.
\footnote{LSDAT implementation will be made public.} 
\end{abstract}
%\begin{keywords}
\begin{figure}
\centering
\includegraphics[width=0.9\linewidth]{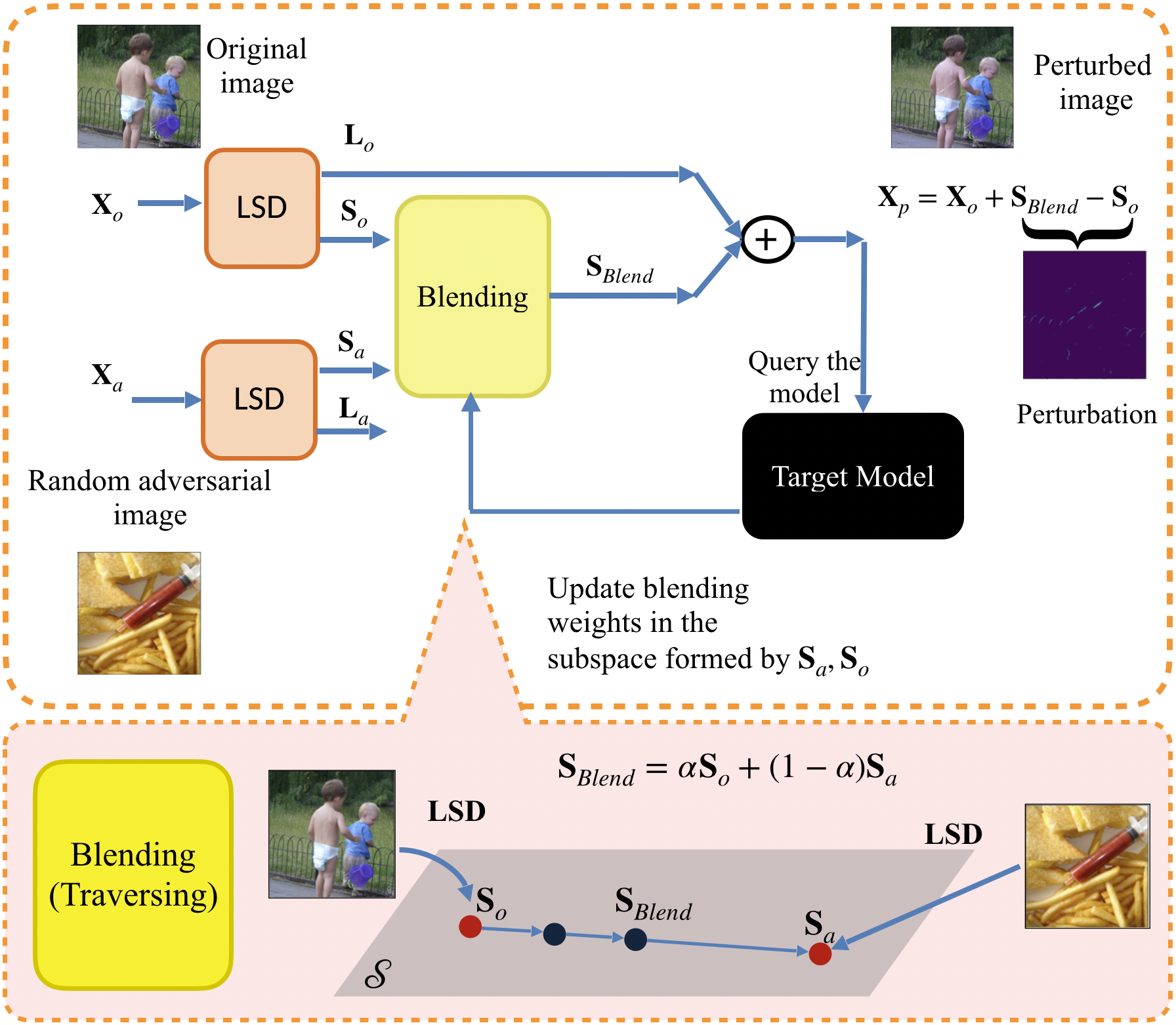}

\caption{ \footnotesize{LSDAT. First, low-rank and sparse decomposition is performed to extract sparse components of the input image and a random initial adversarial sample.
%\NR{in figure with have xadv. clarify and be consistent in using x-inlut, x-target, x-adv,...}.
The sparse components constitute a low-dimensional subspace($\mathcal{S}$) in the image space. %Moreover, the image concepts are encoded in their sparse components. 
%The sparse components contain far fewer non-zero coordinates compared to the original image dimensions and therefore, constitute a low-dimensional subspace. Moreover, the image concepts are encoded in their sparse components. 
%hold the important image concepts under scrutiny within far fewer pixels than the image dimensions. 
Then, the attacker gradually traverses the path between the two sparse components with a pre-determined step size ($\alpha$) and blends the sparse components via a weighted combination. The perturbed image can be obtained by adding the result of blended sparse components to the low-rank component of the input image. 
%As the path lies in the low-dimensional subspace formed by sparse components, LSDAT entails fewer coordinates resulting in query-efficiency and maintaining direct control over sparsity of the attack in the original image domain. 
The traget model is queried with the perturbed image. This iterative process continues until the model is fooled or the query budget gets exhausted. %If the predicted label is unchanged, the attack keeps traversing further, sweeping the sparse path until the attack becomes successful.  %\MS{this is very good figure. However, blending part is not clear; you get  E form M, which is added to E from Eadv, but it shows E becomes Eadv? may be there should be an arrow form Eadv to Eblend.}
}
%\Ashkan{E gradually moves step-wise to be replaced with Eadv. Transformation is a better term compared to blending for confusion.}
}

\end{figure}

%\end{keywords}
%%%%%%%%% BODY TEXT

%%%%%%%%% BODY TEXT
\section{Introduction}\label{sec:intro}

Deep neural networks (DNN) have been employed to yield high accuracy in many image classification related tasks \cite{rawat2017deep}. 
However, despite %obtaining significant accuracy in image classification tasks,
their high accuracy, the possibility of adversarial attacks has cast a shadow over the versatility of deep learning models. This vulnerability to adversarial attacks can lead to catastrophic consequences in applications which are safety-critical such as autonomous driving, robotics, and financial services fraud detection \cite{vorobeychik2018adversarial}. %, to name a few.
Adversarial attacks are applicable to all deep learning based real-world systems such as speech recognition \cite{carlini2016hidden}, speech-to-text conversion \cite{carlini2018audio}, face recognition \cite{sharif2017adversarial}, visual classification \cite{eykholt2018robust}, and other general physical world cases as discussed in \cite{kurakin2016adversarial, athalye2018synthesizing}. In this paper, we focus on the image-classification task.
%Ajmal: the above paragraphs says attacks are bad.
% the next paragraph said (before) so we will launch another attack.
% I put in a sentence to justify the need
Adversarial attacks are mainly studied to gain insights into the inner working of DNNs  %\NR{what do you mean by black-box decisions of DNNs?} 
or to measure their robustness. Adversarial attacks can be launched in a white-box setting, where access to the complete DNN model is available \cite{goodfellow2014explaining,moosavi2016deepfool, carlini2017towards,madry2017towards,papernot2016limitations,ilyas2018prior}, as well as in a black-box setting, where access to only the model output is available \cite{akhtar2018threat}. In the latter case, the DNN model is queried with samples and the outputs are analyzed to optimize the attack. It is critical for adversarial attacks to perform efficiently and with a low-query budget %(i.e. small number of queries to the model under the attack) 
because large number of queries entail computational complexity and can make the target network skeptical of an unusual activity. To this end, we propose Low-Rank and Sparse Decomposition (LSD) for decision based black-box attack; LSDAT for brevity, which is efficient and works under a very low-query budget.
%Ajmal: The contributions read better now.
% how about this
The main contributions of LSDAT are summarized as follows:
%\AM{The contribution statements are long and verbose. They should be succinct.}

\begin{itemize}[noitemsep]
\item We propose an efficient decision-based black-box attack exploiting the low rank and sparse decomposition (LSD) of images to craft sparse adversarial perturbation in a sparse domain. 
\iffalse
The support indexed by sparse components of images form a low-dimensional subspace in the original domain as few non-zero coordinates are associated with sparse components.
\fi
We show that crafting the perturbation in this low-dimensional subspace of sparse representations
%The sparse dimensionality of the sparse component embedded in the image domain
reduces queries and leads to computationally more efficient attack compared to other dimensionality reduction techniques.
%Moreover, we show that sparse components form informative bases representing the image concept.
%We further  provide a comprehensive complexity and convergence rate analysis of \emph{LSDAT} and establish the computational upper-bound for the proposed attack. 
%\item We present an \emph{efficient decision-based black-box attack} (\emph{LSDAT}) that exploits the low rank and sparse decomposition (LSD) of images in crafting adversarial attacks. Benefiting from the properties of LSD, we justify how \emph{LSDAT} achieves query-efficiency compared to the state-of-the-art (SOTA).
%Max-1 query results for LSDAT are also provided showing its efficacy.

%A direction which lieson this subspace and also forms a vertice of`1balls (dueto sparsity and few coordinates involved) isSa−So.  Thegoal is to show that for some initialSa, moving along thepathSa−Sostarting fromXointroduces a viable sparseperturbation which is highly likely to be the most aligneddirection  with  the  shortest  path  to  decision  boundary  (δ)compared to other vertices of the`1ball‖X−Xo‖1=cte

\item
We provide theoretical analyses showing the proposed perturbation direction is highly likely the most aligned sparse perturbation with the shortest path from the original image to the decision boundary for some initial adversarial sample.
%Since
%Therefore, it is effective as it is both sparse and hence norm-constrained, and also likely to cross a decision boundary due to relative alignment with $\delta$. % This is discussed in section \ref{sec:theorem} and highlighted in the geometric illustrative Fig.\ref{fig:geometric2}.
%for the sample to be fooled (original image) and show that proposed perturbation direction 
%generated by traversing the path from the sparse component of original image to that of initial adversarial images
% is highly likely to be aligned with sparse projection projection of direction with the shortest path from original image to the decision boundary of the classifier discussed 

\item We introduce an online learning technique to build prior knowledge from successful attacks. Through sequential attacks, a group of prominent initial adversarial images are organized into 2 levels of class-specific and global dictionaries to be used as candidate initial adversarial samples for upcoming attacks. We empirically demonstrate that the top-1 entry of the global dictionary is one of the \emph{universal} adversarial images and establish theoretical properties for such images.
Exploiting the prior information significantly reduces the average queries.% in the attacking process.  

\end{itemize}

\section{Related Works}

%\ME{We need to cut off the related work }
%The adversarial attacks can be categorized in two broad groups of white box and black box. In white box scenario, the attacker has a full access to the model architecture, parameters and data distribution. Deepfool\cite{moosavi2016deepfool}, FGSM \cite{goodfellow2014explaining}, Carlini \& Wagner attack \cite{carlini2017towards}, Projected Gradient Attack (PGD) \cite {madry2017towards}, Jacobian-based Saliency Map Attack (JSMA) \cite{papernot2016limitations}, Elastic-Net Attack \cite{chen2017ead}, and Adversarial-Bandit Attack \cite{ilyas2018prior} are some of the well known attacks in this category.
The black-box attack scenarios can be categorized based on the attacker's knowledge. The Transfer-based attacks rely on the transferability of adversarial examples among models and exploit substitute models to craft them like attacks proposed in \cite{papernot2017practical,cheng2019improving,huang2019black, liu2016delving}. In this scenario the attacker has access to the data distribution but has no information about the model. Another  category is score-based attack that limits the attacker knowledge only to the model scores such as the class probabilities or logits. %This information allows 
The attacker tries to estimate the gradient of the model from the score through significant number of queries. \cite{bhagoji2018practical,li2020qeba,mohaghegh2020advflow,tu2019autozoom,chen2017zoo,narodytska2017simple,alzantot2019genattack,ilyas2018black} are some of the efficient methods in this group.  
%A query-efficient score based method is presented in \cite{bhagoji2018practical}.
%Other examples of score-based attack method include Gradient Approximation QEBA \cite{li2020qeba}, AdvFlow \cite{mohaghegh2020advflow}, AutoZOOM \cite{tu2019autozoom}, ZOO \cite{chen2017zoo}, LocalSearch \cite{narodytska2017simple}, GenAttack \cite{alzantot2019genattack}, and Query-Limited Partial-Information Attack \cite{ilyas2018black} to name but a few.
Decision-based attacks are the most challenging scenario which narrows the the attackers vision only to the classifier's top-1 hard label output. 
The first work considering decision-based attack was the Boundary Attack (BA) proposed in \cite{brendel2017decision}. BA estimates the boundary and moves along it to minimize the perturbation.   
The Query-Limited Attack \cite{ilyas2018black} leverages a Monte-Carlo approximation approach to approximate the model scores based on the label-setting only and from there on, proceeds with score-based Partial Information Attack as presented earlier. 
%This can be achieved by applying several queries and averaging them to estimate the logits. In addition to inferior accuracy in estimating the logits, this approach contradicts query-efficiency as it requires even more queries to approximate the logits.

Another efficient decision-based method is HopSkipJumpAttack (HJSA) \cite{chen2020hopskipjumpattack} which is based on estimating the gradient direction using top-1 class labels. 
%To further improve the BA approach, Dong, et. al. have introduced 
A natural evolutionary (NE) algorithm has been introduced to update the data covariance matrix after certain queries to reduce the search space from a sphere to an quadratic eclipse characterized by the covariance matrix. 
%As the method proceeds, the empirical covariance matrix of data is updated. Assuming Gaussian prior, the algorithm becomes much faster and more efficient by reducing the search space from sphere to an eclipse characterized by the updated covariance matrix \cite{dong2019efficient}.
A similar approach in updating covariance matrix based on truncated Gaussian distributions is leveraged to adapt Deepfool to the decision-based GeoDA method \cite{rahmati2020geoda}. Natural evolutionary strategies (NES) were first considered by Ilyas, et. al. \cite{ilyas2018black} in designing query-efficient attacks.

Cheng, et. al.~\cite{cheng2018query} model the top-1 label attack as a real-valued optimization problem and use zeroth-order optimization approach to design query-efficient attack using randomized gradient-free method (RGF). Zhao, et. al. \cite{zhao2019design} have proposed ZO-ADMM method where they integrate the alternating direction method of multipliers (ADMM) with zeroth-order (ZO) optimization and Bayesian optimization (BO) to design a query-efficient gradient-free attack. In \cite{cheng2019sign}, the authors extend the optimization based approach and estimate the gradient sign at any direction instead of the gradient itself and introduce Sign-OPT which is more query-efficient compared to OPT. 

Another sign-based method, SIGN-HUNTER \cite{al2019sign} exploits a sign-based gradient approximation rather than magnitude-based to devise a binary black-box optimization. Their method does not rely on hyper-parameter tuning or dimensionality reduction. 
Chen, et. al. have suggested to  randomly flip the signs of
a small number of entries in adversarial perturbations and this way, boost the attacker's performance, specifically in defensive models compared to EA, BA, SignOPT, and HSJA \cite{chen2020boosting}. 
%the authors have proposed an approximation which uses the model decision output to approximate the model logits. Then, the logits can be employed to design the attacks based on the methods which rely on the logits to design the attacks.  
\cite{brunner2019guessing} offers a bias for gradient direction based on a surrogate model. 
RaySearch is another recent approach \cite{chen2020rays}
which advantages over HJSA, and Sign-OPT in efficiency. RaySearch proceeds by finding the closest decision boundary via solving a discrete optimization problem that does not require any zeroth-order gradient estimation.
Dimension reduction based attack techniques are investigated to achieve query efficiency. Sign-OPT-FFT\cite{cheng2019sign}, Bayes Attack \cite{shukla2020hard}, SimBA-DCT\cite{guo2019simple}, QEBA-S, QEBA-F, QEBA-I, and GeoDA-Subspace\cite{rahmati2020geoda} are  which are effective in $\ell_2$ attacks, but in order to guarantee other imperceptibility bounds, they must fetch to the original and transformation domains consecutively, imposing computational burden on their procedure.
Certain methods focus on crafting attacks which are sparse in the image original dimensions such as SparseFool \cite{modas2019sparsefool}, GreedyFool \cite{dong2020greedyfool}, Sparse attack via perturbation factorization \cite{fan2020sparse} (for white-box), and Sparse-RS \cite{croce2019sparse}, CornerSearch \cite{croce2020sparse} and GeoDA (sparse version) \cite{rahmati2020geoda} (for black-box). 

\section{Proposed Method: LSDAT}

The proposed LSDAT method is considered for untargeted black-box adversarial attack. Untargeted attack can be formulated as the following optimization problem:
\begin{equation}
    \underset{\boldsymbol{\delta}}{\min}~~ \Vert {\boldsymbol{\delta}} \Vert_p
    ~~~~ s.t. ~~ \mathcal{C}(\mathbf{X}_0+\boldsymbol{\delta})\neq \mathcal{C}(\mathbf{X}_0),\label{untargeted}
\end{equation}
where $\boldsymbol{\delta}$ denotes the added perturbation. The goal is to minimize the perturbation $\ell_p$-norm such that when applied to the input image $\boldsymbol{x}_0$, the classifier $\mathcal{C}$ is fooled.

Since LSD is one of the main building blocks of the proposed attack, here we briefly introduce it first to be self-contained.
%Our method is based on
LSD is a well-established optimization problem studied in classical machine learning with image and video processing applications including video surveillance, object detection, background subtraction, text extraction, face reconstruction, and manifold learning  \cite{otazo2015low, bouwmans2016handbook, zhou2011godec, liu2015background, azghani2019missing}.
LSD is an image-agnostic and non-data-driven transform which assumes most images can be explained with a low-rank background plus a sparse part.
Mathematically, if the image to be perturbed is denoted by  $\mathbf{X}$, LSD can be formalized as:
\begin{eqnarray}
\underset{\mathbf{L},\mathbf{S}}{\min} & \quad \text{rank}(\mathbf{L})+\lambda\Vert\mathbf{S} \Vert_{0}, 
\quad \text{s.t.}~~\mathbf{X}=\mathbf{L}+\mathbf{S},\label{P1}
\end{eqnarray}
where $\mathbf{L}$ is the low rank and $\mathbf{S}$ is the sparse component. The regularization coefficient $\lambda$ determines the sparsity level of $\mathbf{S}$.
The convex surrogate functions for the rank function and the $\ell_0$-norm are considered to be the trace norm  $\Vert . \Vert_*$ and the $\ell_1$-norm, respectively. Hence, Problem~\eqref{P1} can be cast as a convex optimization problem
\begin{eqnarray}\label{prob:LSD}
\underset{\mathbf{L},\mathbf{S}}{\min} & \quad \Vert\mathbf{L}\Vert_{*}+\lambda\Vert\mathbf{S}\Vert_{1},  
\quad \quad \text{s.t.}~~ \mathbf{X}=\mathbf{L}+\mathbf{S}.
\end{eqnarray}
%\MS{define here norm *}.
%The LSD can be performed via strong toolkits such as Robust Principal Component Analysis (RPCA) or other convex solvers \cite{candes2011robust}.
\iffalse
the low-rank part consists of limited number of large uniform patches. 
\fi
The sparse part consists of far less (in terms of order of magnitude) pixels compared to the original image dimensions; while these reduced data are highly informative and well represent details of the image and are therefore decisive in classification \cite{zhang2011image,zhang2016multi,wright2008robust}.% Figure \ref{fig:LSD-transform} is a sample of LSD transformation. %   To illustrate pictorially, a toy example of original images and their LSD are depicted in Fig. \ref{fig:1}.
Algorithm \ref{alg:LSDAT} summarizes the proposed LSDAT. 
%\AM{How about: Algorithm \ref{alg:LSDAT} summarizes the proposed LSDAT. You have another `Next' a few lines later. This is also a logish paragraph.} 
An initial adversarial sample (with a ground truth label different from the original image) is randomly selected. 
LSD is performed on the original and the adversarial images.
Next, the sparse components of the two images $\mathbf{S}_o$ and $\mathbf{S}_a$ are considered. The union of non-zero pixels in the sparse components form a low-dimensional subspace $\mathcal{S}$ in the original domain. 
There are noticeably fewer coordinates in $\mathcal{S}$ in order for crafting the perturbation which can significantly reduce the query number (which generally scale with the dimensions and why low-dimensional transforms are of interest). Narrowing down the attack vision to $\mathcal{S}$, we propound that for some certain initial adversarial sample $\mathbf{X}_a$, the perturbation direction $\mathbf{S}_a-\mathbf{S}_o$ is the most aligned sparse direction with the shortest path ($\delta$) from $\mathbf{X}_o$ to the decision boundary. This will be theoretically shown in Section \ref{sec:theorem}.
\begin{algorithm}
\caption{LSDAT with one initial adversarial sample} \label{alg:SP}
\iffalse
\algsetup{
linenosize=\small,
linenodelimiter=:
}
\fi
%\fi
\begin{algorithmic}[1]\label{alg:LSDAT}
\footnotesize
\REQUIRE $(\mathbf{X}_o, r)$: Original image and its class, $\mathbf{X}_{a}$: Initial adversarial sample, $\rm{MaxIter}$:  Maximum number of iterations, $\alpha$: sparse traversing rate, $p$: $\ell_p$ constraint type, $\mathcal{T}=$ Imperceptibility constraint budget $\{k,\epsilon,\sigma \}$ \\  \textbf{Output}:$\;\mathbf{X}_{p}$: Perturbed image, $N_Q$: Number attempted queries, $F_{attack}$: Attack success flag \\
\textbf{Initialization:} $\mathbf{X}_{p}\leftarrow\mathbf{0}$,~$N_Q\leftarrow 0$,~ $F_{attack}\leftarrow False$,~ $i\leftarrow1$
\STATE $(\mathbf{L}_o,\mathbf{S}_o) \leftarrow \textbf{LSD}(\mathbf{X}_o)$, $(\mathbf{L}_{a},\mathbf{S}_{a}) \leftarrow \textbf{LSD}(\mathbf{X}_{a})$ 
% \STATE $(\mathbf{L}_{a},\mathbf{S}_{a}) \leftarrow \textbf{LSD}(\mathbf{X}_{a})$ \\ 
%\MS{do you also randomly initialize this?} \Ashkan{written above after initialization, yes}\\
%\STATE $i=1$
\WHILE{$i\leq\rm{MaxIter}$}
\STATE $\mathbf{S}_{i} \leftarrow (\alpha \times i) \mathbf{S}_{a}+(1-\alpha \times i)\mathbf{S}_o$ 
%\MS{are these hard coded numbers?}
\iffalse
\STATE $\mathbf{R} \leftarrow \mathbf{S}_{i}-\mathbf{S}$ 
%\MS{where is E from?}
%\STATE $\mathbf{R}_{imp} \leftarrow \mathbf{R}$ 
\STATE $\mathbf{R}_{imp}  \leftarrow \Pi_{\epsilon} (\mathbf{R}) $
\fi
\STATE $\mathbf{S}_{i} \leftarrow \mathbf{S}_o+\Pi_{p}(\mathbf{S}_i-\mathbf{S}_o,\mathcal{T})$ \label{line_pi}

\STATE $\mathbf{X}_{i} \leftarrow \mathbf{L}_o+\mathbf{S}_{i}$
\STATE $c=\rm{Query}(\mathbf{X}_{i})$
\IF {$c\neq r$}
\STATE $F_{attack} \leftarrow True$ ,~~$\rm{N_Q}\leftarrow$ $i$,~~ $\mathbf{X}_{p} \leftarrow \mathbf{X}_{i} $ 
%\STATE $\rm{N_Q}\leftarrow$ $i$
%\STATE $\mathbf{X}_{p} \leftarrow \mathbf{X}_{i} $
\STATE break
%\RETURN $\mathbf{X}_{pert},\rm{N_Q},F_{attack} $
\ENDIF
\STATE $i=i+1$
\ENDWHILE
\RETURN $\mathbf{X}_{p},\rm{N_Q},F_{attack} $
\end{algorithmic}
\end{algorithm}

%\MS{statements 8-14 unnecessarily complicates the Algorithm with multiple if-then statements; that give an impression this algorithm is pretty rudimentary!;  this  can be simplified to only 3 statements for three norms. Also, can you combine statements 6-8 into one statement?}
% Transform domain representations such as Fast Fourier Transform (FFT) map images to the frequency domain representation. 

\begin{figure}[t!]
\centering
\begin{subfigure}{1\columnwidth}
\centering
    \includegraphics[width=0.9\textwidth]{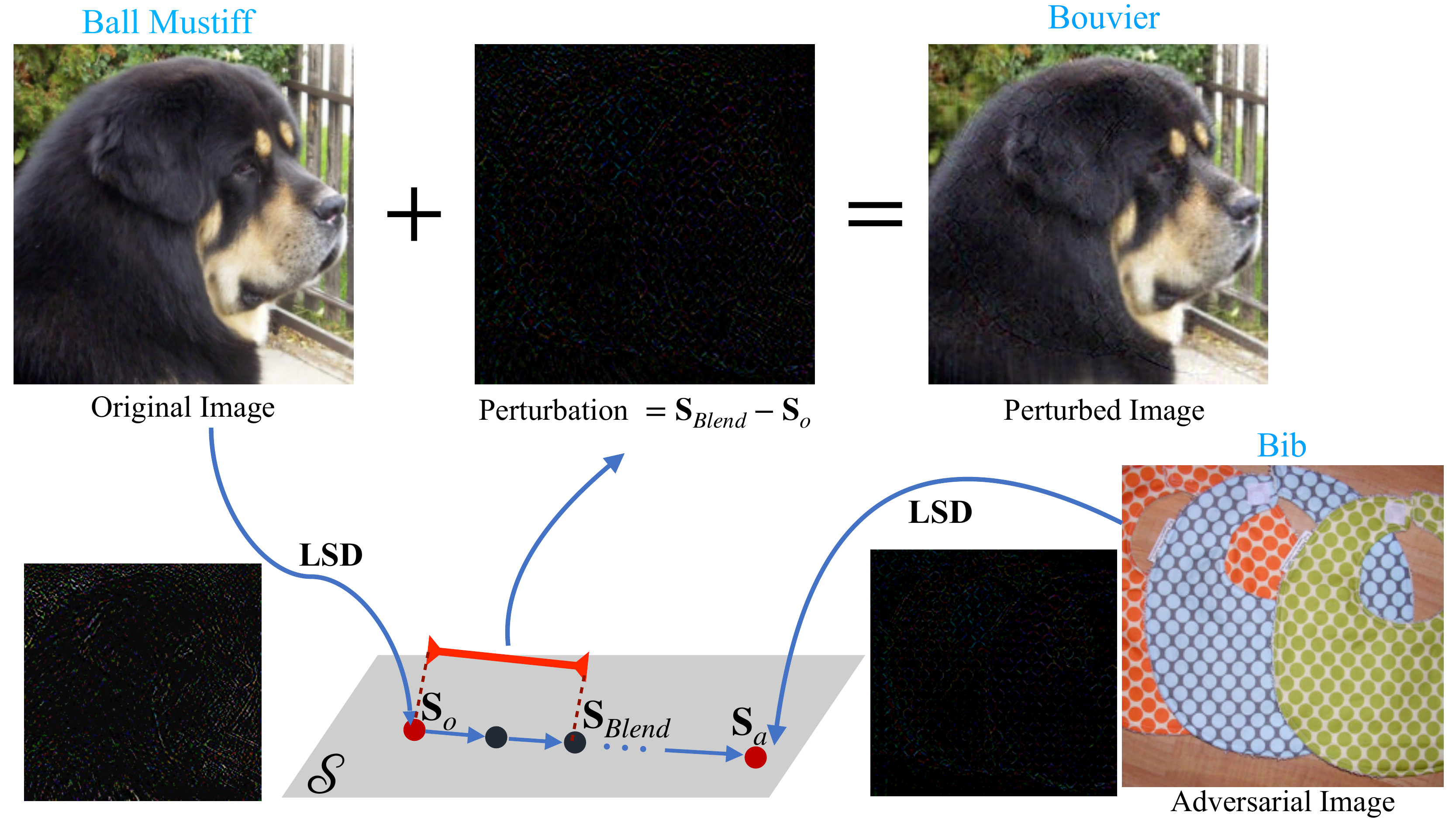}
    
\end{subfigure}
\caption{\footnotesize{
 Illustration of LSDAT. Perturbation lies on the path $\mathbf{S}_a-\mathbf{S}_o$ and is added in a step-wise fashion to $\mathbf{S}_{o}$, which is finally added to $\mathbf{X}_o$. 
}
}\label{fig:illust}

\end{figure}

%Next, we  introduce our proposed LSDAT algorithm (Alg. \ref{alg:LSDAT}). 
 %The key concept governing the algorithm is that the intrinsic dimension of the sparse components are far smaller than the dimension of the original images.
%Hence, a transform from one sparse component to the other does not require as many queries as of the original dimension order.
%In other words, the main %semantic concept and 
%information characterizing the images lie in their sparse component. 
%This means that transforming one sparse component to the other rapidly transforms the %semantic 
%content of the original image to the %semantic
 %the initial adversarial sample.
The attack attempts to induce a new sparse pattern ($\mathbf{S}_a$) into this $\mathbf{X}_p$ which is highly informative of $\mathbf{X}_a$ and suppresses $\mathbf{S}_o$ while traversing from $\mathbf{S}_o$ to $\mathbf{S}_{a}$. %This is carried out in a sparse domain which is lower in complexity compared to the dimensions of the original image. 
It is worth noting that the alteration from one sparse component to another is done via a weighted combination of both. Since the traversing is in a low-dimensional subspace, the semantic transform from $\mathbf{X}_o$ to $\mathbf{X}_a$ is carried out rapidly and within few steps (queries). In other words, small step sizes (perturbations) in the sparse domain leads to more drastic changes in the image concept.  
%The transform is carried out via a weighted sum of the sparse components and hence lies in $\mathbb{S}$ and is therefore sparse. 
%the weighted sum remains sparse and therefore, throughout the procedure, the algorithm moves on a path of sparse matrices which are sparse encodings of the original images.
%Moreover, moving from one sparse component to the other is possible via a small number of queries which provides a fast and efficient transition from the original sample to the perturbed one.
%decisive
In Section \ref{sec:theorem}, we will show how a locally linear classifier (LLC) functions well based on a basis of low rank and sparse components and verify the efficacy of using sparse components as decisive elements in classification. 
Fig. \ref{fig:illust} demonstrates this procedure. %The information transition path is paved on the sparse components with a fast rate thanks to 
%the small intrinsic dimension and semantic transformation. 

After obtaining the perturbation on the specified direction of $\mathbf{S}_a-\mathbf{S}_o$, it is added to
the low-rank component $\mathbf{L}_o$ to form the candidate perturbed image for each query.  
One motivation of preferring LSD to FFT-based methods is that although transform methods like FFT-based or spatial-based approaches optimally represent geometric structure information of images, they cannot extract entire contours and edges accurately, while the LSD can extract the edges and salient parts of an image in an image-agnostic fashion. Side effects, such as pseudo Gibbs phenomenon and false contours are downsides of domain transform-based methods \cite{zhang2016multi}.
%... diret control complexity

To satisfy the imperceptibility constraints in the LSDAT procedure, after obtaining the perturbation $\mathbf{S}_{a}-\mathbf{S}_o$ %\NR{have not defined sblend}
, it is projected on the $\ell_p$-ball depending on the imperceptibility constraint. Projection is denoted by $\Pi_{p}$ in Alg. \ref{alg:LSDAT} line \ref{line_pi}.
we apply appropriate projections depending on the type of imperceptibility constraint ($\ell_p$ norm) denoted with operation $\Pi$ in Alg. \ref{alg:LSDAT} line 6.
\subsection{LSDAT with Imperceptibility Constraints}
Adversarial attacks designed to be confined with certain imperceptibility constraints, to make them feasible, by limiting the norm of the perturbation. $\ell_p$ norm bounds are the prevalent perturbation constraints considered in the literature. In our work, we consider $\ell_0$, $\ell_2$, and $\ell_\infty$ norm bounds on the perturbations.
 \iffalse
In many black-box attack scenarios, the $\ell_\infty$ limit holds for pixel modification. In this part, we adapt the proposed algorithm to restricted $\ell_\infty$ version. 
\fi

\section{Challenge of Selecting Initial Adversarial Sample}

%Exploration and Exploitation for LSDAT \AM{Need a better caption than this one}}
\begin{figure}
\centering
    \includegraphics[width=0.9\linewidth]{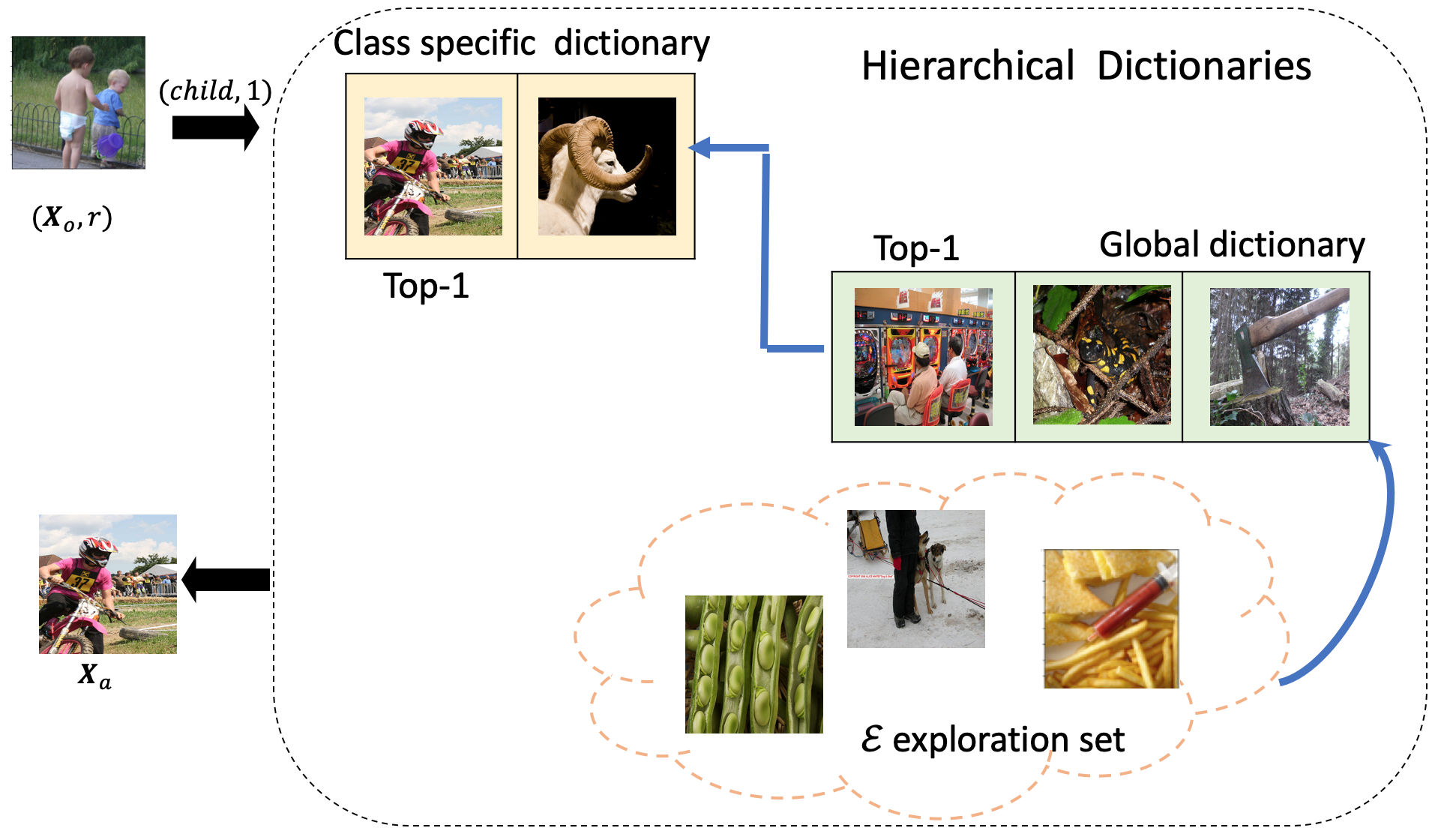}
    %exp.png}
    %\caption{Origina}\label{fig1a}

\caption{\footnotesize Hierarchical dictionary structure for initial adversarial sample provider. At each point the sample is fetched based on the class label r and the index i in the the budget $G$.  
}

\label{fig:exp}
\end{figure}
%\ME{remove this lime}In this section, we introduce the \textit{exploration \& exploitation}
%\AM{ Mention  `direction' or `target'  more specific than exploration and exploitation. use term `Dictionary' in heading and no subheadings. long paragraph.} concept which can be added on top of LSDAT to increase the fooling rate and reduced average queries (AQ).
%As discussed, LSDAT suits low-query budget scenarios. 
%Therefore, it is crucial to take the necessary steps required to increase the fooling rate. In general, the number of queries and the fooling rate are %inversely \NR{they are not inversely proportional, they are proportional} 
%proportional.% forming a trade-off. 
%Hence, the exploration and exploitation concept is put forward which can potentially increase the attack fooling rate without significantly increasing the queries. 
As stated previously, the perturbation direction of interest is set as $\mathbf{S}_a-\mathbf{S}_o$ for some random $\mathbf{S}_a$ to maintain sparsity and small perturbation $\ell_p$ norm.
%In general, there may be several initial sparse patterns as candidates for $\mathbf{S}_a$. 
One random choice may not yield the optimal perturbation direction. 
%Hence, we consider several initial adversarial samples 
%(either from validation set if accessible or otherwise transferred from other sparse pattern \NR{??} of open-set samples) 
%\footnote{Open set refers to the samples which are not generated from the data distribution and the model does not classify them among the known class labels}
Thanks to noticeable query-efficiency of LSDAT, we can explore among %allocate some initial queries to 
several initial adversarial samples. We call this set of random samples the \emph{exploration} set, $\mathcal{E}$. If the initial adversarial image budget of an attack per sample is G, the set $\mathcal{E}$ is gathered such that $|\mathcal{E}|=G$. To launch an attack to the image $\mathbf{X}_o$, we consider one sample at the time, drawn from $\mathcal{E}$ as the initial adversarial sample and apply Alg. \ref{alg:LSDAT} to it. Figure \ref{fig:exp} Note that, if the attack is successful at any point, the rest of the exploration set will not be attempted. Thus, the number of queries is %upper bounded by
$(j\times MaxIter)+N_{Q}$, where $j$ is the number of unsuccessful initial adversarial attempts and $N_{Q}$ is the number of queries used in the successful attack.% However, practically, far fewer AQ is required to craft an adversarial image. 
\iffalse
We will see in the theoretical section why it is important to search in the exploration set to find an initial sample which forms an $\epsilon$-net with $\mathbf{X}_o$ covering the decision boundary in a proximity of $\mathbf{X}_o$.
\fi
%All extra initial adversarial samples contain a sparse component. Without loss of generality, assume the sparsity level is less than $k$. The sparse components of different adversarial samples lie in the set $\mathbf{B}_0=\{\mathbf{S}| \Vert \mathbf{S} \Vert_0 \leq k\}$. We call this set the exploration set, from which a suitable initial adversarial sample is picked to facilitate the entire attack procedure. 

%The goal is to find an optimal sparse component from this set which is closest to the $\mathbf{S}$ (the sparse component of the input image) in $\ell_P$ norm. 

%As the $\ell_0$ ball is non-convex, one can consider the $\ell_1$ ball $\mathcal{B}_1=\{\mathbf{S}| \Vert \mathbf{S} \Vert_1 \leq thr\}$ \MS{for some threshold $thr$}  to represent the sparsity and benefit from convexity. The exploration set $\mathcal{B}_1$ is a convex set and finding the closest sparse component $\mathbf{S}_*$ to $\mathbf{E}_{init}$ is an $\ell_p$ projection problem. An augmented exploration set enhances the probability of locating a good initial adversarial sample. Hence, during LSDAT procedure, the distance between the sparse components becomes less, leading to what we coin as

\subsection{Online Learning with Hierarchical Dictionaries}

%As discussed, exploration set facilitates fooling the model by introducing an expansive sparse patterns.
Black-box attack scenarios can be categorized into isolated and non-isolated ones. In non-isolated attacks, where the goal is to attack a set of images rather than a single one, the attacker can build a prior knowledge through the attacking process by learning the set of elite samples which are universal in fooling the previously attacked input images.
These prominent samples are organized into a hierarchy of \emph{class-specific} and \emph{global} dictionaries. The intuition is that if one sample serves as a good initial adversarial image for multiple images of a specific class; for instance the class cat; it is also very likely to be a good initial adversarial point for the other instances of that class. If the class specific dictionary does not have any entry or the number of entries is limited, the best initial adversarial points are those which have successfully fooled other classes so far. All dictionaries' entries are always ranked based on their score, which is defined as the number of successful attacks for that image as an initial adversarial image up to now.  
Our proposed dictionary-based attack exploits the previous good initial adversarial samples to launch a new attack with fewer queries. In
a new attack on an image $\mathbf{X}$ with label $r$ and the initial adversarial image budget of G, first the entries of class-specific dictionary for class $r$ are selected one after the other as the initial adversarial image. If none of them leads to an adversarial counterpart for $\mathbf{X}_o$ and the budget G is not met, the remaining initial images are picked from the global dictionary and if exhausted, from random sampling (Figure \ref{fig:exp}). If the attack is successful, we update the dictionaries accordingly, i.e. update the score for the dictionaries entries or adding a new item to them. As the attacking process continues, the dictionaries become richer to the point that top-1 entry of the global dictionary contributes significantly in successful attacks, we refer to this sample as 
\emph{universal} adversarial initial image. The properties of such images are investigated theoretically in the next section.
\iffalse
To this end, we equip LSDAT to score the elements in the exploration set based on their success in fooling the model. Candidates are ranked based on their score in the dictionary and are pinged from the dictionary in order to achieve highest efficiency. 

In classification task, a class-specific dictionary of ... inja 

In isolated attacks no online exploration is possible. Hence, the attacker can use offline images transferable to the model and train susbtitute networks  
When a sample in the exploration set is useful in fooling an input image
the \textit{exploitation} property. The result is smaller perturbation norm as well as fewer number of queries for fixed step size. 
The key concept behind the exploration \& exploitation scenario is that a random initial adversarial image may give the attacker a relatively inappropriate initial point, subsequently leading to larger $\ell_p$ norm perturbation on the image. Further explanation on how to implement this is provided in the experiments section. The exploration set concept is shown in Fig. \ref{fig:exp}.
%\section{Introducing universal target samples}
In this section, we introduce our method in creating a dictionary of samples which are useful in universally fooling the source input images. To this end, we consider samples which are successful in fooling the input image and score them to obtain a dictionary of samples with success score reflecting how many times the input samples have been fooled by the mentioned samples.
\fi

\section{Theoretical Analysis}\label{sec:theorem}

Now, we establish the theoretical analysis for LSDAT. 
First, it is worth noting that most of decision-based attack methods in the literature rely on estimating the decision boundary, the gradient, and in some instances the second derivative. Obtaining an estimation the aforementioned necessitates  
certain number of queries to the model, which contravenes the query-limited assumption. Our method is designed for extremely limited query regime. Thus, estimating the decision boundary or the gradient are avoided in LSDAT procedure and it is adapted for extremely low-query budgets. (In experimental section, we include "max-1" query attack results for LSDAT to manifest LSDAT functionality under extremely query-limited scenarios. This is despite the fact that most SOTA methods are dysfunctional for very limited budgets.
%We want to put forward why the perturbation  $\mathbf{S}_o-\mathbf{S}_a$ is a viable sparse perturbation to fool the model. 
\iffalse
\begin{figure}%{0.9\columnwidth}
\centering
    \includegraphics[width=1.\linewidth]{figs/geo.png}
    \caption{\footnotesize{ Geometric illustration of LSDAT. The attempt is to show that among sparse directions, $\mathbf{S}_a-\mathbf{S}_o$ is likely to be the most aligned one with the shortest path to decision boundary $\delta$, and therefore is likely to cross the decision boundary as well as maintaining perturbation norm efficiency.}}\label{fig:geometric2}
    
\end{figure}
\fi
\iffalse
\begin{figure}%{0.9\columnwidth}
\centering
    \includegraphics[width=1\linewidth]{figs/intersection.png}
    \caption{Original}\label{fig1a}
\end{figure}
\fi
We assume the original and the adversarial image are decomposed as $\mathbf{X}_o=\mathbf{L}_o+\mathbf{S}_o$, $\mathbf{X}_a=\mathbf{L}_a+\mathbf{S}_a$,  where $\mathbf{L}_o, \mathbf{S}_o$ and $\mathbf{L}_a, \mathbf{S}_a$ denote the low rank and sparse components of the original and the adversarial images, respectively. The goal is to show that $
\mathbf{S}_a-\mathbf{S}_o$ is a viable sparse perturbation centered around $\mathbf{X}_o$ that can fool the model. 
Before delving into the analysis, we elaborate on the geometric interpretation of LSDAT functionality, as depicted in Fig. \ref{fig:geometric2}. 
 It is known that an $\ell_1$-ball centered at $\mathbf{X}_o$ has sharp corners (vertices). If one gradually enlarges congruent $\ell_1$-balls centered at $\mathbf{X}_o$, it is highly likely that one of them intersects the decision boundary at one of its sharp corners which is a well-known property of $\ell_1$ contours. 
 Certain $\ell_1$-balls centered at $\mathbf{X_o}$ also intersect with the subspace spanned by $\mathbf{S}_o$ and $\mathbf{S}_a$ denoted as $\mathcal{S}$. The intersection can be formulated as $\sum_{i}w_{1i} |s_{oi} | + \sum_{j}w_{2j}
 |s_{aj}| = cte$ 
 %\NR{what is $S_{oi},...$}.
 where $s_{oi}$ and $s_{aj}$ are the $i$ and $j $ element of $\mathbf{S}_o$ and $\mathbf{S}_a$ respectively.
 A specific direction which lies on $\mathcal{S}$ and also forms a vertice for one $\ell_1$-ball (due to sparsity 
 %\NR{remove->and few coordinates involved}) 
 is $\mathbf{S}_a-\mathbf{S}_o$.
 %One of such co is obtained by traversing $\mathbf{S}_a-\mathbf{S}_o$.
 The goal is to show that for some initial $\mathbf{S}_a$, traversing the path $\mathbf{S}_a-\mathbf{S}_o$ starting from $\mathbf{X}_o$ introduces a viable sparse perturbation which is highly likely to be the most aligned sparse direction with the shortest path to decision boundary ($\delta$) compared to other vertices of the $\ell_1$-ball $\Vert \mathbf{X}-\mathbf{X}_o\Vert_1=cte$. Therefore, 
 %\NR{what it refers to? mention that}
 it is both sparse and hence norm-constrained, and also likely to cross a decision boundary due to relative alignment with $\delta$. The described concept can be visually observed in Fig. \ref{fig:geometric2}. 
 Now, we delve into mathematical analyses. 
 First, we assume the decision boundary can be locally linearized in part of an $\epsilon$-net
 %\footnote{$\epsilon$-net is a set that is both $\epsilon$-packing and $\epsilon$-covering .} 
 covered by $\mathbf{X}_o$ and some initial $\mathbf{X}_a$ in the exploration set $\mathcal{E}$ (for larger $\|\mathcal{E}\|$, LSDAT is more likely to find such $\epsilon$-net), 
 %falls within $\epsilon$ distance from the decision boundary, 
 where $\epsilon$ is a small value.
 %The model decision boundary can be estimated with a \emph {locally linear classifier} (LLC) near the decision boundary. 
 In general, considering there are $P$ nearest samples in an $\epsilon$-net covering local decision boundary, a locally linear classifier (LLC) can be estimated using regression on a basis composed of low-rank and sparse components
 %$\{\mathbf{L}_i,\mathbf{S}_i\}_{i=1}^P$, 
%a basis of $P$ elements 
$\mathcal{B}=\{\mathbf{L}_i, \mathbf{S}_i\}_{i=1}^P$. The local regression weights can be used for classification (fed to a linear SVM for instance).

It is shown in \cite{zhang2011image} that LLC trained on the basis formed by LSD components of nearest samples yield favorable classifier with small generalization error. An LLC is governed by certain regression weights near each sample. This leads to a nonlinear classifier in general as different parameters are to be utilized for different local regions. Considering an $\epsilon$-net covering decision boundary consisting of $\mathbf{X}_o$, the desired LLC can be obtained as follows:
\begin{align}
    \underset{\mathbf{c}_p}{\min}~~\sum_{p=1}^{P}~~ \Vert \mathbf{X}_p-\mathcal{B}\mathbf{c}_p \Vert^2 +\lambda \Vert \mathbf{d}_p \odot \mathbf{\mathbf{c}_p}\Vert^2,
\end{align}\label{eq:LLC}
where $\mathbf{d}_p$ is defined as $\mathbf{d}_p=\exp{(\frac{dist(\mathbf{X}_p,\mathcal{B})}{\sigma}})$. $\sigma$ determines the decay rate for locality, and $dist(\mathbf{X}_p,\mathcal{B})$ is the distance between $\mathbf{X}_p$ and basis elements in $\mathcal{B}$.
%As our method leverages several initial candidates for successfully fooling the model, without loss of generality, we assume some initial adversarial sample $\mathbf{X}_a$ and the original image $\mathbf{X}_a$ are the only two samples lying in $\epsilon$ neighborhood of the local decision boundary. 
Therefore, any sample in this $\epsilon$-net can be written using the low rank and sparse basis expansion as
\begin{align}
    \mathbf{X}_p=\mathcal{B}\mathbf{c}_p= \bar{\mathcal{B}}(\epsilon)\bar{\mathbf{c}}_p+\mathcal{B}(\epsilon)\mathbf{t}, 
\end{align}
where $\mathcal{B}(\epsilon)=[\mathbf{L}_o,\mathbf{S}_o,\mathbf{L}_a,\mathbf{S}_a]$, $\bar{\mathcal{B}}(\epsilon)=\mathcal{B} \backslash \mathcal{B}(\epsilon)$, and $\bar{\mathbf{c}}_p, \mathbf{t}$ are split components of $\mathbf{c}_p$ indexing $\bar{\mathcal{B}}(\epsilon)$ and $\bar{\mathcal{B}}(\epsilon)$, respectively.
  %$\bar{\mathbf{c}}_p=\mathbf{c}_p \backslash$ \mathbf{t}.
The LLC prioritizes the components based on the distance from the samples. We have assumed the $\epsilon$-net is covered by $\mathbf{X}_o$ and $\mathbf{X}_a$. Thus, they are dominant terms in LLC, and for some $\tau(\epsilon, \sigma)$ which is increasing w.r.t $\sigma$ and $\epsilon$,
%\begin{equation}
$\Vert \mathbf{t} \Vert > \big(1-\tau(\epsilon,\sigma)\big) \mathbf{c}_p.$
%
%\end{equation} 
This means, 
\begin{equation}\Vert \mathbf{X}_p-\mathcal{B}(\epsilon)\mathbf{t}\Vert =\Vert \bar{\mathcal {B}}(\epsilon)\bar{\mathbf{c}}_p \Vert \leq \tau(\epsilon,\sigma) \Vert \bar{\mathcal {B}}(\epsilon)\Vert_{op} \Vert \bar{\mathbf{c}}_p\Vert \end{equation}
For small $\sigma$ and $\epsilon$ values, $\tau$ becomes small and the dominant components of $\mathbf{c}_p$ index $\mathcal{B}(\epsilon)$.
Taking the latter into account,
%\begin{align}
 $   \mathbf{X}_p= \mathcal{B}(\epsilon)\mathbf{t} +\mathcal{O}(\tau) \approx \mathcal{B}(\epsilon)\mathbf{t}$.
%\end{align}

We are specifically interested in some point $\mathbf{X}_p$ (as the perturbed image) that lies in the $\epsilon$-net and while maintaining the sparsest perturbation form $\mathbf{X}_o$, fools the model. 
%Let the perturbed signal be $\mathbf{X}_p$.
Let $\mathbf{t}=[t_1,t_2,t_3,t_4]$. The perturbation $\mathbf{X}_p-\mathbf{X}_o$ can be written as,
\begin{align}
\mathcal{B}(\epsilon)\mathbf{t}=(t_1-1)\mathbf{L}_o+(t_2-1)\mathbf{S}_o+t_3\mathbf{L}_a+t_4\mathbf{S}_a
\end{align}
It is desired that the perturbation 1- be sparse (or $\ell_p$ bounded) as much as possible, 2- be aligned with the side information fooling direction $\mathbf{X}_a-\mathbf{X}_o$ as much as possible, and 3- the perturbed image $\mathbf{X}_p$ be close to $\mathbf{X_a}$ in the $\epsilon$-net as much as possible so as to cross the boundary and fool the model. Therefore, to find the desired perturbation, the following optimization over $\mathbf{t}$ is suggested:
\begin{align}
    & \underset{[t_1,t_2,t_3,t_4]}{\min}~~ \mu \underbrace{\Vert (t_1-1)\mathbf{L}_o+(t_2-1)\mathbf{S}_o+t_3\mathbf{L}_a+t_4\mathbf{S}_a \Vert _1}_{\text{sparsity measure}}+ \nonumber \\ 
    & \lambda \underbrace{\theta\bigg((t_1-1)\mathbf{L}_o+(t_2-1)\mathbf{S}_o+t_3\mathbf{L}_a+t_4\mathbf{S}_a, \mathbf{X}_a-\mathbf{X}_o\bigg)}_{\text{alignment with the difference direction~}\mathbf{L}_a+\mathbf{S}_a-\mathbf{L}_o-\mathbf{S}_o}+ \nonumber \\ 
    & \underbrace{\Vert t_1\mathbf{L}_o+t_2\mathbf{S}_o+(t_3-1)\mathbf{L}_a+(t_4-1)\mathbf{S}_a \Vert_2}_{\text{distance of $\mathbf{X}_p$ to the adversarial sample}}\label{prob:pert}
\end{align}
%The solution to P \eqref{prob:pert} depends on regularization coefficients $\lambda$ and $\mu$.
%When $\lambda \rightarrow \infty$, $\mathbxf{t}=[0,0,1,1]$ is a trivial solution ($\mathbf{X}_a$). When $\lambda < \infty$ 
When $\mu$ is large enough (which is a reasonable assumption enforcing restricted perturbation norm), coefficients of $\mathbf{L}_o$ and $\mathbf{L}_a$ tend to 0 in the $\ell_1$ regularized term promoting sparsity because these are largely non-sparse terms compared to $\mathbf{S}_o$ and $\mathbf{S}_a$ (similar to sparse group lasso \cite{simon2013sparse}). This leads to $t_1=1, t_3=0$. 
%Next, we assume $\mathbf{L}_o-\mathbf{L}_a$ is orthogonal to $\mathbf{S}_o-\mathbf{S}_a$ which is
Therefore, the sparsity-constrained term shrinks to $\Vert (t_2-1)\mathbf{S}_o+ t_4 \mathbf{S}_a \Vert_1$. Assuming orthogonality of linear combination of $\mathbf{S}_o$ and $\mathbf{S}_a$ to $\mathbf{L}_a-\mathbf{L}_o$ \footnote{This is a reasonable assumption on the basis $\mathcal{B}$.}
, the third term (distance of $\mathbf{X}_p$ to $\mathbf{X}_a$) can be written as $\Vert t_1 \mathbf{L}_o + (t_3-1)\mathbf{L}_a\Vert_2$+$\Vert t_2\mathbf{S}_o + (t_4-1)\mathbf{S}_a\Vert_2$. As stated, $t_1$ and $t_3$ values are forced by large $\mu$.    
The compromise between the controllable expression in the third term $\Vert t_2\mathbf{S}_o + (t_4-1)\mathbf{S}_a\Vert_2$ and the sparsity regularizer, i.e., $\Vert t_2\mathbf{S}_o + (t_4-1)\mathbf{S}_a\Vert_2$+$\mu \Vert (t_2-1)\mathbf{S}_o+ t_4 \mathbf{S}_a \Vert_1$ determines the blending (linear combination) weights in $\mathcal{S}$.
The sparsity regulaizer is minimized for $t_2=1, t_4=0$ (trivial solution, and the $\ell_2$ term minimization yields $t_2=0, t_4=1$. 
Therefore, when there is a compromise of both (large $\mu<\infty$), the solutions lie somewhere on the sparse perturbation vector ($\mathbf{0},\mathbf{S}_a-\mathbf{S}_o$). 
The second term in loss function also works in favor of the $\ell_2$ term, weighing the balance of solutions towards $t_2=0,t_4=1$. (It is worth noting that as $\mu$ is large, the $\ell_1$ term is dominant. Thus, solution path tends to a linear interpolator between ($\mathbf{0},\mathbf{S}_a-\mathbf{S}_o$). )
It is worth noting that as the $\epsilon$-net setting may not be realized and therefore the basis expansion on sparse components and the aforementioned optimization is not valid. This may lead to larger perturbation norm to cross decision boundary corresponding to other classes. The point is although the direction is not optimal (to save for numerous queries required to estimate the gradient and decision boundary, a sparse perturbation which is likely to align with it is introduced.), thanks to query-efficiency we can attempt many samples from $\mathcal{E}$ as far as the query budget allows. This gives rise to the probability of crossing some decision boundary even if the adversarial sample is not optimal via LSDAT procedure. Moreover, the perturbation may cross another decision boundary before crossing the desired one which is in this case favorable to the method's performance, making LSDAT suitable for untargeted attack.

As mentioned, there may exists some universal samples in a dictionary of elite samples which are globally capable of fooling the model for input samples from diverse classes.
Although the path $\mathbf{S}_a-\mathbf{S}_o$ has been shown to be the most aligned (best sparse approximation) sparse direction with $\delta$, yet this alignment can vary depending on existence of the $\epsilon$-net and the angle between $\delta$ and $\mathbf{S}_a-\mathbf{S}_o$. The angle depends on how sparse the $\delta$ is itself. Theoretically, they are universally most aligned if
%In other words, for some $\mathbf{X}_a$ whose texture (low-rank) is insignificant and it is almost captured in its sparse component, the $\delta$ direction is aligned with $t1_\mathbf{S}$ if for all $\mathbf{X}_o$,
%$\delta$ lies on $\mathcal{S}$ characterized for each $\mathbf{X}_o$.
$\delta$ is close to its sparse approximation $\mathbf{S}_a-\mathbf{S}_o$ as much as possible. As stated before, $\delta$ is expressed in the basis $\mathcal{B}(\epsilon)$ and will be the sparsest if in the representation $(t_1-1)\mathbf{L}_o+(t_2-1)\mathbf{S}_o+t_3\mathbf{L}_a+t_4\mathbf{S}_a$, $t_3=0$ or to put it differently,
the sample itself is almost its sparse component and its texture (low-rank component) is negligible or very small. 

\footnote{***\emph{As we care about sparse attack, we have only provided analysis for $\ell_1$ loss on $\delta$ in P\eqref{prob:pert}. We have provided $\ell_2$ and $\ell_\infty$ attacks adapting the proposed sparse perturbation to other non $\ell_1$ scenarios. One can
interchangeably, adapt the first loss to $\ell_2$  and include terms incorporating $\mathbf{L}_o$ and $\mathbf{L}_a$. The reason we suffice to the sparse version is simplicity by traversing the proposed path in a sparse domain involving few pixels. ***}}

It is worth noting that although the method is designed for the sparse ($\ell_1$) attack, as $\ell_1$ and $\ell_2$ norms of a vector are cohered, the $\ell_2$ also is likely to shrink which makes LSDAT to be applicable for $\ell_2$ attacks as well. Although, the presented sparse perturbation is not necessarily same as the minimial $\ell_2$ distance to the decision boundary, the partial alignment contains information of the minimal perturbation and makes the sparse perturbation also likely to cross the decision boundary. 
\begin{table*}[t]
\footnotesize
\centering
\begin{tabular}{l|l|c|c|c|c|c} 
\toprule
&\multicolumn{3}{c|}{Resnet-50}&\multicolumn{3}{c}{VGG-bn}\\
%\hline
        &\quad\quad$\epsilon=5$ & $\epsilon=10$ & $\epsilon=20$ &  $\epsilon=5$ & $\epsilon=10$ & $\epsilon=20$\\
 Method & FR \quad\quad\quad AQ & FR \quad\quad\quad AQ &  FR \quad\quad\quad AQ&FR \quad\quad\quad AQ & FR \quad\quad\quad AQ &  FR \quad\quad\quad AQ\\
\midrule
\midrule
  BA\cite{brendel2017decision} & 8.52 \quad\quad\quad 666.5 & 15.39 \quad\quad\quad 577.9 & 26.97 \quad\quad\quad 538.1& 11.23 \quad\quad\quad 626.3 & 21.27 \quad\quad\quad 547.6 & 39.37\quad\quad\quad 503.2\\
   OPT Attack\cite{cheng2018query} &  7.64 \quad\quad\quad 777.4 &  15.84 \quad\quad\quad 737.2& 32.53 \quad\quad\quad 757.9& 11.09 \quad\quad\quad 736.6 & 21.79 \quad\quad\quad 658.9 & 43.86\quad\quad\quad 718.7 \\
  HJSA\cite{chen2020hopskipjumpattack} &  6.99 \quad\quad\quad 904.3 &  14.76 \quad\quad\quad 887.1 & 28.37 \quad\quad\quad 876.8 & 10.30 \quad\quad\quad 893.2 & 21.53 \quad\quad\quad 898.2 & 40.82 \quad\quad\quad 892.6\\
 Sign-OPT\cite{cheng2019sign} &  7.46 \quad\quad\quad 777.4 & 15.84 \quad\quad\quad 737.1 &32.53 \quad\quad\quad 757.9& 19.81 \quad\quad\quad 841.1 & 35.8 \quad\quad\quad 843.7 & 60.63 \quad\quad\quad 857.7 \\
   %Sign-OPT-FFT \cite{cheng2019sign} &  13.99 \quad\quad\quad 919.1 &  29.26 \quad\quad\quad 906.3 &  55.97\quad\quad\quad902.8& 21.4 \quad\quad\quad 916.8 & 42.93 \quad\quad\quad 910.9 & 70.81 \quad\quad\quad  907.9 \\
   Bayes Attack\cite{shukla2020hard} &  20.10 \quad\quad\quad 64.2 &  37.15 \quad\quad\quad 64.1 &  66.67\quad\quad\quad 54.97&24.04 \quad\quad\quad 69.8 & 43.46\quad\quad\quad 76.5 & 71.99 \quad\quad\quad 48.9\\\midrule
   LSDAT(R) & \textit{23.40} \quad\quad\quad  \textit{53.9} & \textit{47.6} \quad\quad\quad \textit{41.5} & \textit{75.20} \quad\quad\quad \textit{35.6}& \textit{30.20} \quad\quad\quad \textit{58.8} & \textit{55.6} \quad\quad\quad \textit{43.8} & \textit{81.00}\quad\quad\quad \textit{33.9}\\
   LSDAT(D) &\textbf{25.40} \quad\quad\quad \textbf{35.2} & \textbf{47.6} \quad\quad\quad \textbf{39.4} & \textbf{76.80} \quad\quad\quad \textbf{21.50} & \textbf{32.80} \quad\quad\quad \textbf{36.4} & \textbf{56.80} \quad\quad\quad \textbf{32.9} & \textbf{82.40}\quad\quad\quad \textbf{15.2}\\
  \bottomrule
\end{tabular} %*************************

\caption{\footnotesize Comparison of different $\ell_2$ attack methods performance on ImageNet based on various $\ell_2$ ball constraint $\epsilon$ for Effect of hyper parameters ). FR and AQ stand for fooling rate and average query respectively. In LSDAT(x), x="R" represents random samples, x="D" stands for Dictionary base. Best performances are in bold. }
\label{tab:ell2}

\end{table*}
\begin{figure}%{0.9\columnwidth}
\centering
    \includegraphics[width=1.\linewidth]{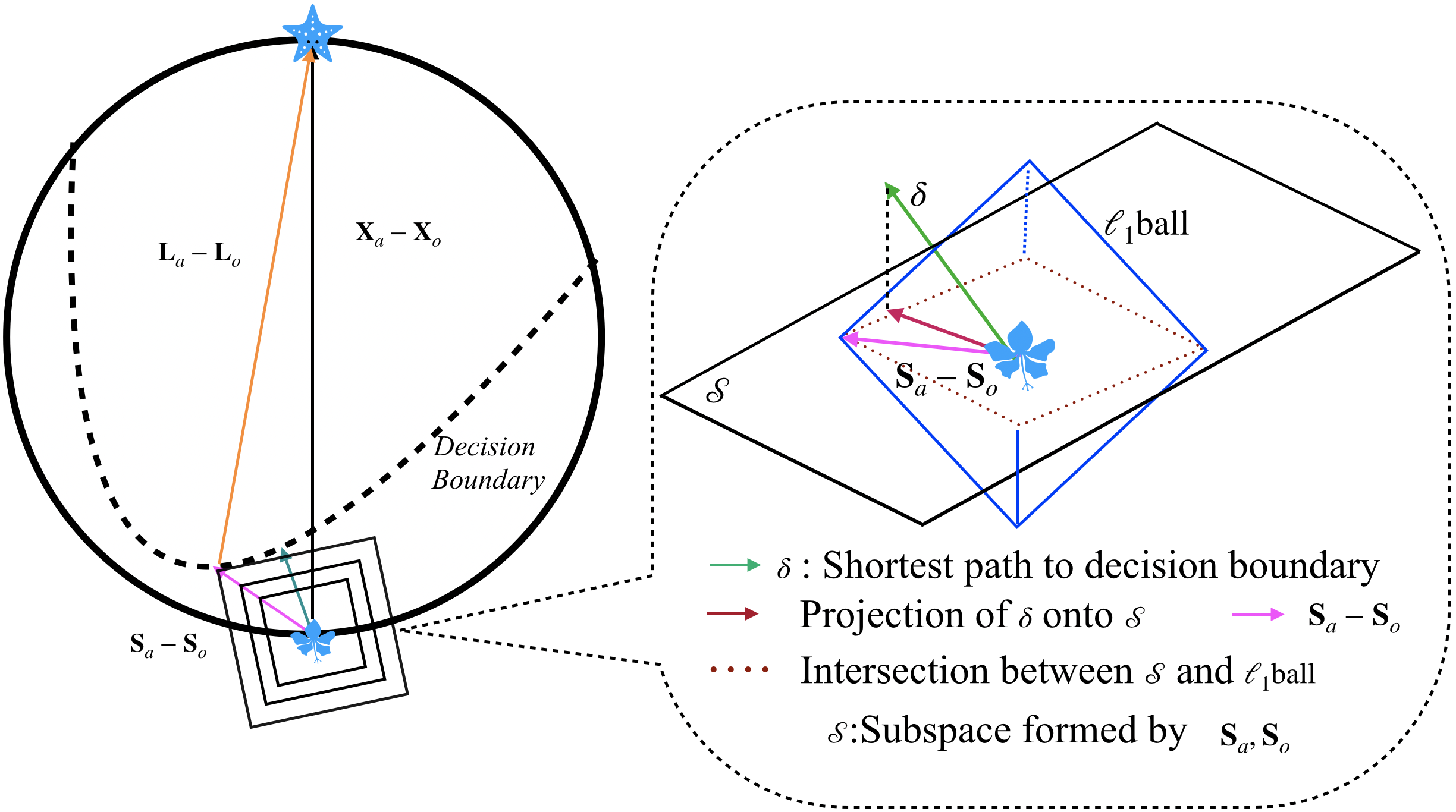}
    \caption{\footnotesize{ Geometric illustration of LSDAT. The attempt is to show that among sparse directions, $\mathbf{S}_a-\mathbf{S}_o$ is likely to be the most aligned one with the shortest path to decision boundary $\delta$, and therefore is likely to cross the decision boundary as well as maintaining perturbation norm efficiency.}}\label{fig:geometric2}
\end{figure}
\section{Complexity and Convergence Rate Analysis}

%In this section, we provide complexity analysis for the proposed method. %Ajmal: of course we do as the title says
A general fact is that the number of queries scale with the image dimensions as each coordinate can play a role in fooling the model.
To remedy the curse of dimensionality in crafting adversarial perturbation for high-dimensional data, domain transforms are applied to the original image in order to design the perturbation in a low-dimensional space. 
%As cases in point, QEBA-S, QEBA-F, and QEBA-I use spatial, low-frequency, and intrinsic component subspace transforms to adapt QEBA to low-dimensional attack, and significantly reduce the required quer
%\cite{li2020qeba}. ies to achieve the same fooling performance. ***
FFT-based methods such as QEBA-F \cite{li2020qeba}, Bayes-Attack \cite{shukla2020hard} are the most query-efficient attack method to the best of our knowledge. Although 
reducing the required queries, performing low-dimensional transforms and their inverse impose extra computational burden per each query.  
%FFT imposes an $\mathcal{O}(n\log_2 n)$ query-wise computational complexity on the proposed method. 

The complexity of the FFT-based methods (for $m=n$) are dominated by %(in the best case which happens when each low-dimensional query-wise attack is crafted simply via one iteration) 
$\mathcal{O}(2N\times t \times n^2\log_2 (n))$, where $N$ is the number of queries, $n$ denotea original dimension, $t$ is the iterations per query for FFT and its inverse IFFT.
%\footnote{In general, dimension-reduction transforms such as PCA have complexity  $\mathcal{O}(mn\min\{m,n\})$). FFT is privileged over PCA due to its implementation structure.} 
Although increasing the efficiency, such transforms come at the cost of increased query-wise complexity.  
%(For RGB images, ensemble of channels dimension can be considered). 
Additionally, FFT-based methods do not obtain control on non- $\ell_2$ (such as $\ell_0$ or $\ell_\infty$) perturbation constraints in the transform domain. This mandates applying extra transforms to perform clipping, thresholding, and projections back in the original domain in order for satisfying such imperceptibility constraints. Extra transforms come at the cost of more computational burden. While LSDAT directly maneuvers the image in the original domain obtaining direct control on such constraints. 

%This adds to the cost of more queries and contradicts the low-query budget assumption. 
%Although, we have provided such projection options for LSDAT, but setting the coefficient $\lambda$ in P. \ref{prob:LSD} to a proper value, sparsity of the crafted attack is guaranteed in our design. The thresholding is provided in the algorithm to make it instance independent.
%To refrain from notation prolixity, we assume $m$ is equal to $n$, which is a reasonable assumption for most image datasets. 
% Assuming $m=n$, the FFT based approach yields the complexity of $\mathcal{O}(4tNn^2\log_2 n)$ required as both FFT and IFFT must be evaluated. 
%\footnote{There are other crafting operations of $\mathcal{O}(n^2)$ such as perturbation projections which are dominated by the $2Nn^2\log_2 n$.}
%To be more precise in terms of computational complexity, the concept \textit{arithmetic complexity} is used to show the number of multiplications and additions.
%The fastest FFT evaluation has been shown to require $\frac{34}{9}n \log_2 n$ operations \cite{lundy2007new}. Hence in general, the number of operations required for FFT-based methods, (disregarding $\mathcal{O}(n^2)$ details such as crafting details (perturbation projections)) are $\sim 4\times \frac{34}{9}Nn^2\log(n)$.
On the contrary,
LSDAT merits over such methods as it only applies a one-time initial LSD for initial samples attempted from the exploration set. Next, it applies summations on sparse components in the original domain. As the sparse coding and the summation all happen in the original domain, there is no additional transform-related computational burden per query in LSDAT.
%LSDAT does not need to apply any transform to find the candidate perturbed image for each query. 
 The most efficient computational complexity corresponding to RPCA is obtained by accelerated alternating projections algorithm (IRCUR) \cite{cai2020rapid} which is (for $m=n$) $\mathcal{O}(Gnr^2log_2^2(n)log_2 (\frac{1}{\epsilon}))$,
 where $r$ is the rank of the low-rank component, and $\epsilon$ is the accuracy of the low-rank component (appearing as the number LSD solver algorithm iterations), and $G$ is the number of explored samples in the exploration set $\mathcal{E}$. It immediately follows that the proposed are less complex compared to transform based methods with a factor of $\frac{log(n)}{n}$ which plays important role in high-dimensional setting.

Now, we provide convergence rate analysis for the proposed algorithm. Since the LSDAT attack relies on the weighted combination of sparse component of original image and that of adversarial image  for fast semantic transform of the original image, and these sparse components are obtained at the early stage of the algorithm, the convergence rate only depends on how fast the weighted combination phase, sweeps the path between sparse components. 
%As mentioned earlier, LSDAT gradually sweeps the path between the sparse component of nitial adversarial image and that of the input image in a sparse domain transform through a weighted combination (fast semantic transform, i.e. the initial adversarial image sparse concept is infused and the input sparse parts is faded). Since our method does not search for coordinates (pixels) to perturb and the sparse components under attention are obtained at the early stage of the algorithm, the convergence rate only depends on how fast the weighted combination phase, sweeps the path between sparse components. 
Tuning the step size may not be a priority for $\ell_0$ and $\ell_\infty$ attacks as their perturbation metric does not depend on the distance swept between sparse parts. The choice of step-size, however, affects the perturbation norm for $\ell_2$ constraint.
Thus, the convergence rate of a successful attack only depends on the chosen step size.
%Assuming the attack succeeds fooling, the convergence rate depends on the step size chosen. 
which can be set based on  
how much the perturbation budget. 
%In general, one can simply hard code the step-wise sweeping between sparse parts.
The convergence rate is then $\frac{1}{\alpha}$, where $\alpha$ as also used in Alg.~\ref{alg:LSDAT}, is the modification rate between sparse parts. Large $\alpha$s lead to faster convergence.
% As the proposed algorithm deals with transforming \MS{(I did not see this)} a sparse matrix to another one, it is intuitively expected that the number of queries depend on the sparsity number of target and initial samples. 

\section{Experimental Set-up and Results}

%\begin{figure*}
%    \centering
%    \includegraphics[width=0.7\linewidth]{figs/dic_dist_qvs_top1.png}
%    \caption{Caption}
%    \label{fig:my_label}
%\end{figure*}
\begin{figure*}
\centering
\begin{subfigure}{0.28\linewidth}
    \includegraphics[width=\textwidth]{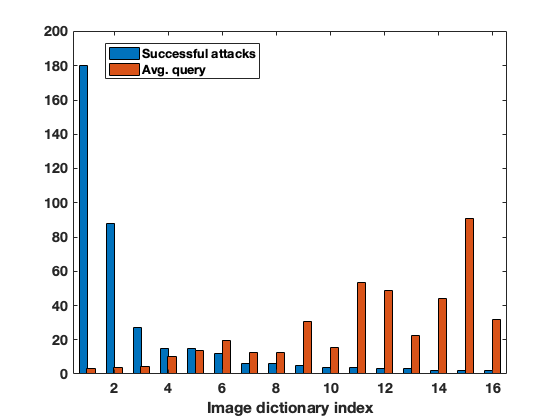}
    \label{fig:gsd}
\end{subfigure}
\begin{subfigure}{0.28\linewidth}
    \includegraphics[width=\textwidth]{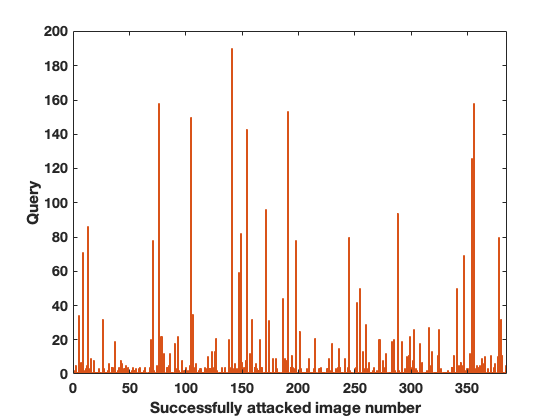}
    %\caption{}
    \label{fig:qvs}
\end{subfigure}
\begin{subfigure}{0.18\linewidth}
    \includegraphics[width=0.8\textwidth]{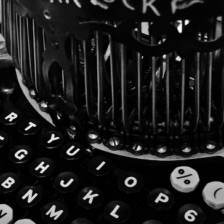}
    \label{fig:gts}%\caption{}
    
\end{subfigure}
\begin{subfigure}{0.18\linewidth}
    \includegraphics[width=0.8\textwidth]{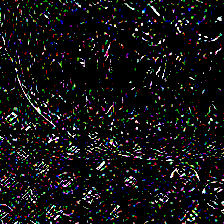}
    %\caption{}
    \label{fig:sgts}
\end{subfigure}
 
\caption{\footnotesize{From left to right,(1) Successful attacks (blue bars) and average query distribution of global dictionary samples with success rate$>1$. The orange bar represents the average query per successful attack. (2) The number of queries for each successful attack. (3) The top-1 image of the global dictionary. (4) The sparse component of the top-1 image which is scaled for the sake of visibility.     
}}

\label{fig:dictionary}
%\vspace{-mm}
\end{figure*}

In this section, we present a comprehensive set of experiments to demonstrate the efficacy of the proposed LSDAT attack in degrading the performance of well-trained image classifiers for ImageNet.
%In separate subsections, we discuss the results of experiments implemented for ImageNEt and Cifar-10 datasets. 
%\subsection{ImageNet}
We apply LSDAT to attack two ImageNet pre-trained models\footnote{https://pytorch.org/docs/stable/torchvision/models.html}, namely ResNet-50 \cite{he2016deep} and VGG16 \cite{simonyan2014very} on the set $\mathcal{D}$, created by gathering correctly classified samples from the ImageNet validation set. 
We increase the step size in LSDAT implementations for $\ell_0$ and $\ell_\infty$ scenarios as it does not affect the corresponding constraints.
Fast convergence of LSDAT with large step size allows us to expand the exploration set $\mathcal{E}$ to increase the fooling rate. 
%As discussed in the convergence rate section, our algorithm is expected to converge fast when the step-size is large. With this in mind, exploration set can grow larger to contain much more initial adversarial points and the attacker has a larger set of initial samples to choose from. Bounding the query budget for each initial sample to $2-3$ queries for a large step-size, the scope considered in the exploration set grows larger.
From now on, we abbreviate Average Queries and Fooling Rate with AQ and FR, respectively.
In the following experiments, FR is defined based on the number of misclassified samples divided by the number of samples in $\mathcal{D}$. The reported number of queries is averaged on all successful attack instances.
%In this set of experiments, we have put our proposed method into test for
%In this section, we provide the experiments on two well-known vision and image datasets, Cifar-10 and ImageNet. 
We compare the performance of LSDAT based on AQ, FR, and perturbation norm with state-of-the-art (SOTA) methods.
%We apply the LSDAT versions on $500$ initially correctly classified validation images of ImageNet and compared the number of queries, success rate and the perturbation norm to the state-of-the-art. 
%In this section, we assume whether the attacker has limited access to validation set of adversarial samples. Hence, it is provided with limited access to process the validation data and explore the validation set in order to select a warm-start. The result in investing for a limited number of queries at the beginning, however, after exploring the validation set an initial warm-start sample is selected such that the final number of queries is less on average owing to the initial exploration property.

\noindent \textbf{LSDAT with $\ell_2$ constraint}: For a fair comparison in $\ell_2$ attack scenario, we use $1000$ samples for the set $\mathcal{D}$ and the initial adversarial samples budget is set to 100. For LSDAT(R), we select the initial samples randomly from validation set. Note that initial sample set varies for each image to be attacked. In LSDAT(D), we exploit the prior information by first selecting initial samples from the class specific (if exists) and then the global dictionary. If the initial adversarial sample budget is not met, the remaining samples are selected randomly. It is worth noting that we only add an image to the initial adversarial set, if the current images could not lead to a successful attack so far.  
 The comparison of performance with the SOTA methods is presented in Table \ref{tab:ell2}. The LSDAT attack consistently outperforms all methods with a significant drop in AQ. LSDAT(R) leads to on average 28.7$\%$ and 29.6$\%$  
 reduction in AQ while it improves the fooling rate by 17.38$\%$ and 31.16$\%$ for ResNet-50 and VGG models, respectively compared to Bayes attack\cite{shukla2020hard}. Applying LSDAT(D), the attack further improves the FR while reducing the AQ  by 47.6$\%$ for Resnet-50 and 57$\%$ for VGG compared to Bayes attack. The AQ gain is one order of magnitude compared to other methods.
%\subsection{Cifar-10}
\begin{table}[t]
\centering
\footnotesize
\begin{tabular}{c|c|c} 
\toprule
         & ResNet-50 &  VGG16-bn\\
  Method & FR \quad\quad\quad\quad AQ & FR \quad\quad\quad\quad AQ \\
\midrule
\midrule
  OPT Attack\cite{cheng2018query} &  5.73  \quad\quad\quad246.3 &  7.53  \quad\quad\quad 251.2\\
 Sign-OPT\cite{cheng2019sign} &  10.31  \quad\quad\quad 660.4 &  15.85  \quad\quad\quad 666.6 \\
  Bayes Attack\cite{shukla2020hard} &  67.48  \quad\quad\quad 45.9 &\textbf{78.47} \quad\quad\quad \textbf{33.7} \\\midrule
  LSDAT(R)  &\textbf{70.00}\quad\quad\quad 31.3 &76.20  \quad\quad\quad 43.2 \\
  LSDAT(D) & 69.40\quad\quad\quad \textbf{29.4} & 74.80  \quad\quad\quad 37.3 \\
 \bottomrule
\end{tabular}

\caption{\footnotesize Comparison of performance of LSDAT $L_{\infty}$ attack for $\sigma=0.05$ with STOA methods.}% \Ashkan{why VGG worse?}}
\label{tab:Linf1}

\end{table}
\noindent\textbf{Universal Adversarial Sparse Image}: We also analyze the effectiveness of dictionary in reducing the AQ and finding the universal adversarial sparse image. To this end we apply LSDAT(D) attack with $\ell_{2}$ constraint of $\epsilon=20$ on a set with $|\mathcal{D}|=500$. The initial sample budget is 100 with \textit{MaxIter=2} per sample. This setting leads to 384 successful attacks.  Figure \ref{fig:dictionary}-left shows the distribution of dictionary samples that bring about at least 2 successful attacks in blue bars along with the AQ per attack for each image in orange bars. Manifestly, the top-1 image in the global dictionary gives rise to 46.8$\%$ of successful attacks with as low as  3.27 query per attack and the top-5 samples are responsible for 86.7$\%$ of successes with average of 7.11 query per attack. These findings support the existence of a universal sample with a sparse component whose difference to the input sparse component is highly likely to align with the shortest path from $\mathbf{X}_o$ to the decision boundary ($\delta$). The top-1 image and its sparse component are illustrated in the last 2 images of Figure \ref{fig:dictionary} respectively. Clearly, the sparse component contains most of the details including keys while the background(texture) is black. 
%To be more precise, in temrs of arithmetic complexity, the best case FFT implementation needs $\frac{34}{9}nlog_2(n)$ operations. Putting all together, for LSDAT to dominate FFT-based, we should have:
%\begin{equation}
    %Gnr^2log_2^2(n)log(\frac{1}{robust \epsilon}) < \frac{34}{9}\times 4Nn^2log_2(n)
%\end{equation}
%which reduces to $Gr^2log_2(n)log(\frac{1}{\epsilon}$
 
\noindent\textbf{LSDAT with $\ell_{\infty}$ or $\ell_{0}$ constraint}: In case of $\ell_{\infty}$ attack, the attack setting is the same as $\ell_{2}$ constrained attack. Table \ref{tab:Linf1} summarizes the performance comparison of LSDAT with SOTA methods when the $\ell_\infty$ perturbation bound is $\sigma=0.05$. The proposed attack, consistently outperforms all methods for ResNet-50 architecture while it achieves similar results as the runner-up method for the VGG16-bn architecture. We believe the lower performance on VGG16-bn roots in the ability of the model in extracting richer features by considering both local and global spatial information, compared to ResNet which makes the attack more difficult.% So the effect of local perturbation caused by LSDAT is compensated by global information and      

Finally, we compare LSDAT attack in $\ell_{0}$ scenario with GeoDA \cite{rahmati2020geoda} in Table \ref{tab:l0}. GeoDA achieves the best FR with limited query budget compared to other sparse attacks such as Sparse-RS \cite{croce2020sparse,croce2019sparse}. Also, Bayes Attack \cite{shukla2020hard} is not the first choice to apply for $\ell_{0}$ constraint as it suits $\ell_2$ and the sparsity level is less controllable in frequency domain due to FFT transformation mandating computational burden and further queries. %and CornerSearch \cite{croce2019sparse}.
%It is worth noting that, as mentioned in Section\ref{sec:intro}, Bayes Attack \cite{shukla2020hard} is not the first choice to apply for $\ell_{0}$ constraint since the sparsity level is less controllable in frequency domain due to FFT transformation. %Moreover, authors in \cite{shukla2020hard} have not provided $\ell_0$ implementations. 
To have a fair comparison with GeoDA, the set $\mathcal{D}$ contains 500 samples with the initial adversarial sample budget $G=100$. %The performance comparison is reported in the Table \ref{tab:l0}. 
Both, LSDAT(R) and LSDAT(D) significantly outperform GeoDA and improve the AQ by at least one order of magnitude . Also, the superiority of LSDAT is clear in highly imperceptible $\ell_{0}$ attacks when only $0.5\%-1\%$ of coordinates are perturbed.
FR is improved up to $21.2\%$ and $17\%$ by perturbing only $0.5\%$ and $1\%$  coordinates respectively, setting the state-of-the-art performance for the imperceptible $\ell_{0}$ attacks.

\begin{table}[t]
\centering
\footnotesize
\begin{tabular}{c|c|c |c} 
\toprule
 \multicolumn{4}{c}{Method}    \\
%\midrule
\midrule
   \multirow{2}{*}{P$\%$}&  GeoDA\cite{rahmati2020geoda} & LSDAT(R) & LSDAT(R)  \\
   &  FR\quad\quad AQ & FR\quad\quad AQ & FR\quad\quad AQ  \\\midrule%\cmidrule{2-4}
 4.29 &88.44\quad 500& 85.20\quad12.6& \textbf{90.00}\quad\textbf{8.3}\\% 87.00 &\textbf{90.4}\\
 3.05 & 82.30\quad500& 80.20\quad15.3& \textbf{83.40}\quad\textbf{8.6}\\%\textbf{85.20} &\textbf{82.6}\\
 2.36 &75.20\quad500& 76.80\quad15.3& \textbf{80.10}\quad\textbf{10.0}\\%73.00 & 112.1  \\
 1.00& 47.00\quad 500& 60.60\quad 17.6&\textbf{64.00}\quad\textbf{12.2} \\
 0.50 &30.00\quad 500 & 49.80 \quad24.2&\textbf{51.20}\quad \textbf{17.2} \\% \textbf{45.00} & 
    \bottomrule
\end{tabular}

\caption{\footnotesize Comparison of LSDAT $\ell_{0}$ attack performance to Resnet-50 model under various perturbation rates (P$\%$)  with GeoDA. In LSDAT(x), x="R" represents random initial adversarial samples, x="D" stands for Dictionary base.}

\label{tab:l0}
\end{table}

\noindent\textbf{Attacking Adversarialy Robust Models}:
We also evaluate the effectiveness of LSDAT against adversarially robust models. To this end, we consider the method proposed by \cite{wong2020fast} for fast adversarial training that leads to a robust Resnet-50 classifier on ImageNet with $43\%$ robust accuracy on PGD attacks.
The result of comparison of LSDAT with GeoDA with various perturbation rate for $\ell_{0}$ constraint attacks are reported in Table \ref{tab:l0-defence}. While LSDAT(x) achieve higher FR with significantly lower AQ compared to GeoDA, we noticed that LSDAT(R) slightly outperforms LSDAT(D) in terms of FR. This phenomena is expected as the adversarial training changes the shape of decision boundary and makes the dictionary entries with small score less reliable as an inertial adversarial images. This necessitates finding a balance factor between exploration set and exploiting dictionary. We postpone this study to our future works. 

\noindent\textbf{Pure Black-Box Attack}:
The most challenging type of black-box attacks is known as pure black box attack in which \emph{only one query} is allowed to launch an attack. %othe attacker has no information about the model, and data distribution and \emph{only one query} is allowed to launch an attack.
We evaluated the performance of the LSDAT in this scenario. LSDAT(R) achieves FR of 25.8$\%$ and 33.6$\%$ on Resnet-50 and VGG respectively for $\ell_{2}$ constraint of $\epsilon=20$. With \emph{only} $1\%$ perturbation on $\ell_{0}$  constraint attacks, the FR=24.4$\%$ for Resnet and FR=21$\%$ for VGG can be obtained.  
Finally on $\ell_{\infty}$ attack with constraint of $\sigma=0.05$ the FR is 26$\%$ and 25.4$\%$ on ResNet and VGG, respectively. Note that other decision-based black box attacks are not applicable in this threat models as they demand more than one query to estimate the decision boundary. For instance, GeoDA\cite{rahmati2020geoda} requires at least 10 queries to obtain average $\ell_{2}$ distance of 39.4 which is as twice as LSDAT with a single query.

\begin{table}[t]
\centering
\footnotesize
\begin{tabular}{c|c c c} 
\toprule
 \multicolumn{4}{c}{Method}    \\
%\midrule
\midrule
   \multirow{2}{*}{P$\%$}&  GeoDA\cite{rahmati2020geoda}  & LSDAT(R) & LSDAT(D)  \\
   &  FR\quad\quad AQ & FR\quad\quad AQ & FR\quad\quad AQ  \\\midrule%\cmidrule{2-4}
 4.29 &71.30\quad 500&73.00\quad19.71& \textbf{73.00}\quad\textbf{9.8}\\% 87.00 &\textbf{90.4}\\
 3.05 & 60.10\quad500& \textbf{65.00}\quad20.3& 62.20\quad\textbf{13.1}\\%\textbf{85.20} &\textbf{82.6}\\
 2.36 &54.70\quad500& \textbf{60.00}\quad24.0& 58.10\quad\textbf{10.8}\\%73.00 & 112.1  \\
 1.00& 36.80\quad 500& \textbf{44.00}\quad 27.2&43.00\quad\textbf{18.7} \\
 0.50 &22.60\quad 500 & \textbf{30.00}\quad32.2& 26.20\quad \textbf{21.1} \\% \textbf{45.00} & 
    \bottomrule
\end{tabular}

\caption{\footnotesize {\footnotesize Comparison of LSDAT $\ell_{0}$ attack performance to an adversarialy robust Resnet-50 model under various perturbation rates (P$\%$) with GeoDA. In LSDAT(x), x="R" represents random initial adversarial samples, x="D" stands for Dictionary base.}}

\label{tab:l0-defence}
\end{table}

\section{Concluding Remarks}

A query-efficient decision-based adversarial attack (LSDAT)  is introduced based on low-rank and sparse decomposition. The method is suitable for very limited query budgets and is of low complexity compared to SOTA. LSDAT is also effective in fooling rate dominating the SOTA in performance as verified through diverse set of experiments. 
LSDAT finds a sparse perturbation which is likely to be aligned with the sparse approximation of the shortest path from input sample to the decision boundary. We show the path lies on the path connecting original and some adversarial sparse components. Theoretical analyses buttresses LSDAT performance in fooling.
%As few pixels entail the image information in the sparse component and the number of queries is relevant to the effective dimension of image to be fooled, the proposed method acts as a query-efficient attack. 
Moreover, LSDAT offers better control over imperceptibility constraints in the original domain and less complexity compared to SOTA as it does not apply consecutive transforms and their inverse. 
%unlike other dimension reduction techniques which craft the perturbation in the transform domain and therefore, lose control on image $\ell_p$ properties while crafting, LSDAT finds the perturbation in the original domain to have control on imperceptibility constraints and has complexity and faster convergence rate.
Experiments on the well-known ImageNet dataset shows query efficiency and fooling rate superiority of LSDAT compared to SOTA. %including their subspace attack (low-dimensional) version.

{\small
\bibliographystyle{ieee_fullname}
\bibliography{arxiv}
}
\end{document}